\documentclass[twoside,twocolumn,10pt]{article}
\usepackage[backend=bibtex, style=numeric, natbib=true]{biblatex}
\usepackage{eucal}
\usepackage{amsfonts}
\usepackage{multirow, booktabs}
\usepackage{stfloats}
\usepackage{afterpage}
\usepackage{wscg}           % includes a number of other packages (e.g., myalgorithm)
\RequirePackage{ifpdf}
\ifpdf
 \RequirePackage[pdftex]{graphicx}
 \RequirePackage[pdftex]{color}
\else
 \RequirePackage[dvips,draft]{graphicx}
 \RequirePackage[dvips]{color}
\fi
\usepackage{nopageno}       % no page numbers at all; uncomment for final version
\title{Detect and Correct: A Selective Noise Correction Method for Learning with Noisy Labels}

\author{
\parbox{0.22\textwidth}{\centering
Yuval Grinberg\\[1mm]
Bar Ilan University, Immunai Inc.\\
juvs2k@gmail.com
}
\hspace{0.05\textwidth}
\parbox{0.22\textwidth}{\centering
Nimrod Harel\\[1mm]
Tel Aviv University\\
nimrodharel@mail.tau.ac.il
}
\hspace{0.05\textwidth}
\parbox{0.22\textwidth}{\centering
Jacob Goldberger\\[1mm]
Bar Ilan University\\
jacob.goldberger@biu.ac.il
}
\hspace{0.05\textwidth}
\parbox{0.22\textwidth}{\centering
Ofir Lindenbaum\\[1mm]
Bar Ilan University\\
ofirlin@gmail.com
}
}

\usepackage{url}
\makeatletter
\def\Uslash{\mathbin{\mathchar`\/}\@ifnextchar{/}{\kern-.15em}{}}
\g@addto@macro\UrlSpecials{\do \/ {\Uslash}}
\def\Ucolon{\mathbin{\mathchar`:}\@ifnextchar{/}{\kern-.1em}{}}
\g@addto@macro\UrlSpecials{\do : {\Ucolon}}
\makeatother

\begin{document}

\twocolumn[{\csname @twocolumnfalse\endcsname

\maketitle  % full width title

\begin{abstract}
\noindent
Falsely annotated samples, also known as noisy labels, can significantly harm the performance of deep learning models. Two main approaches for learning with noisy labels are global noise estimation and data filtering. Global noise estimation approximates the noise across the entire dataset using a noise transition matrix, but it can unnecessarily adjust correct labels, leaving room for local improvements. Data filtering, on the other hand, discards potentially noisy samples but risks losing valuable data.
Our method identifies potentially noisy samples based on their loss distribution. We then apply a selection process to separate noisy and clean samples and learn a noise transition matrix to correct the loss for noisy samples while leaving the clean data unaffected, thereby improving the training process. Our approach ensures robust learning and enhanced model performance by preserving valuable information from noisy samples and refining the correction process. We applied our method to standard image datasets (MNIST, CIFAR-10, and CIFAR-100) and a biological scRNA-seq cell-type annotation dataset. We observed a significant improvement in model accuracy and robustness compared to traditional methods.
\end{abstract}

\subsection*{Keywords}
Noisy Labels, Transition Matrix, Image Classification, Single-cell RNA Sequencing, Cell Annotation, Label Noise Correction, Selective Correction
\vspace*{1.0\baselineskip}
}]

%%%%%%%%%%%%%%%%%%%%%%%%%%%%%%%%%%%%%%%%%%%%%%%%%%%%%%%%%%%%%%%%%%%%%%%%%%%%%

\section{Introduction}
\label{introduction}
The performance of deep learning models is heavily influenced by the quality of their training data \cite{song2022learning}. Noisy labels- incorrect or misleading- pose a significant challenge, as they can substantially degrade these models' accuracy and generalization capabilities. This issue is pervasive across various applications, including image classification \cite{ALGAN2021106771} \cite{wang2019symmetric}, natural language processing \cite{wu2023noisywikihowbenchmarklearningrealworld} \cite{wu2018learning}, and speech recognition \cite{9763412} \cite{ravanelli2017contaminated}, where training data often contains annotation errors or other noise sources.

Numerous methods have been developed to address the challenge of noisy labels. Traditional approaches include loss correction techniques that adjust the loss function to account for label noise, often relying on a noise transition matrix \cite{NEURIPS2022_98f8c89a}. This matrix models the probability of label noise and adjusts the learning process accordingly. However, accurately estimating the transition matrix is not only challenging \cite{pmlr-v202-liu23g} but also introduces an inherent trade-off. Even with a perfectly estimated matrix, both correct and incorrect labels are affected, as the adjustments made to account for label noise can inadvertently impact clean samples, leading to performance issues.

Another avenue of research has focused on developing methods to dynamically detect and filter noisy labels. Techniques such as Co-teaching \cite{NEURIPS2018_a19744e2} and MentorNet \cite{jiang2018mentornet}, which involve training multiple networks simultaneously to filter out noisy examples based on their disagreement, have shown promise. Similarly, the O2U-Net \cite{huang2019o2unet}, introduced in 2019, analyzes the loss behavior of each sample during the training process to identify noisy labels, demonstrating significant improvements in detecting noisy data. More recent advancements have focused on refining and integrating these techniques with sophisticated models to further enhance their effectiveness in filtering noisy labels.

All loss correction strategies based on the noise transition matrix offer a global representation of the label noise distribution \cite{xia2019anchorpointsreallyindispensable}. While these methods capture general noise patterns, they often miss outliers due to an inherent trade-off between correcting mislabeled data and preserving true labels. Conversely, noise detection approaches that discard potentially noisy samples risk losing valuable training data \cite{wei2021learning}, including both correctly labeled examples and those for which the loss could have been corrected.

Building on these advancements, we propose D\&C-Net (Detect and Correct), a hybrid approach combining noise detection with targeted correction. Our method aims to identify potentially noisy samples and uses them to estimate the transition matrix more accurately. This matrix is then applied specifically to these noisy samples, enhancing the training process without altering clean data. 

A notable approach for handling noisy labels is DivideMix \cite{li2020dividemix}, which uses a Gaussian Mixture Model (GMM) to separate clean and noisy samples, treating the latter as unlabeled data for semi-supervised learning. While effective, this approach discards mislabeled samples rather than correcting them. In contrast, our method integrates noise detection with targeted correction, using identified noisy samples to estimate a transition matrix that is selectively applied. This preserves more useful training signals while mitigating label noise, striking a better balance between robustness and data efficiency.

Our contribution goes beyond simply ensuring robust learning from clean data; it incorporates both noise detection and targeted correction strategies to mitigate the impact of noisy labels. Additionally, we introduce a loss distribution analysis using a GMM \cite{muthen1999finite} to distinguish between noisy and clean labels more effectively. We evaluate our method on several benchmark datasets and compare its performance against state-of-the-art approaches, demonstrating its effectiveness in enhancing model performance under noisy conditions.

\section{Problem Formulation}
\label{background}

Consider a dataset \( \mathcal{D} = \{(x_n, y_n)\}_{n=1}^N \), where \( x_n \in \mathbb{R}^d \) represents the feature vector, and \( y_n \in \{1,2, \dots, C\} \) is the true (clean) label. However, in the presence of label noise, the observed labels \( \tilde{y}_n \) (noisy labels) may differ from the true labels \( y_n \). We denote $\mathrm{Y}$ and $\tilde{\mathrm{Y}}$ as the random variables representing the true and noisy labels, respectively. Consequently, we define the noisy dataset as \( \tilde{\mathcal{D}} = \{(x_n, \tilde{y}_n)\}_{n=1}^N \). This noise process can be modeled using an unknown noise transition matrix \( T \in [0,1]^{C \times C} \), where each element \( T_{k,l} \) represents the probability that the true label \( \mathrm{Y} = l \) is flipped to the noisy label \( \tilde{\mathrm{Y}} = k \):
\begin{equation*}
    T_{l,k} = \Pr(\tilde{\mathrm{Y}} = k \mid \mathrm{Y} = l), \quad \forall l,k \in \{1,2,\dots,C\}.
\end{equation*}
By definition, \( T \) is a row-stochastic matrix, meaning that each row sums to one:
\begin{equation*}
\sum_{k=1}^{C} T_{l,k} = 1, \quad \forall l \in \{1,2,\dots,C\}.
\end{equation*}
Given a noisy dataset \( \tilde{\mathcal{D}} = \{(x_n, \tilde{y}_n)\}_{n=1}^N \), our objective is to train a classifier \( f_{\theta}(x) \) (a neural network with parameters \( \theta \)) that can accurately predict the true labels despite the presence of label noise and the unknown transition matrix.

\section{Method}
\label{method}

Our approach (D\&C-Net) separates the training process into two distinct stages, pre-training and training, to effectively tackle the challenge of noisy labels. The pre-training stage is a local process focused on detecting noisy labels within the dataset. This stage operates at a granular level, examining individual samples by analyzing their loss trajectories during training \cite{yi2019probabilistic}. By identifying and removing these noisy labels, we set the stage for a more precise and effective learning process.

\begin{figure*}[htb!]
\begin{minipage}[b]{1.\linewidth}
  \centering
  \centerline{\includegraphics[width=16cm]{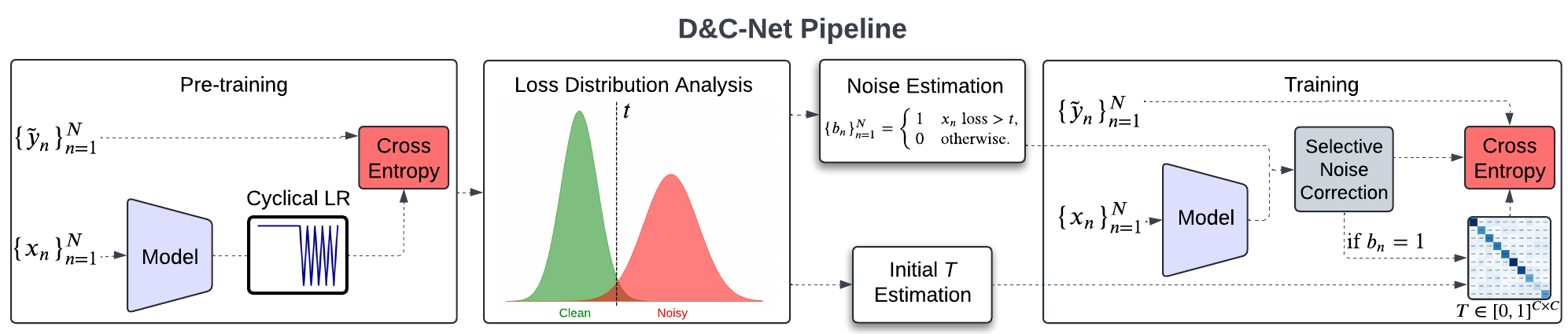}}
\end{minipage}
\caption{This figure illustrates the D\&C-Net method for handling noisy labels. Initially, in the pre-training phase, a cyclical learning rate is employed to estimate two Gaussian distributions: one representing the clean loss distribution and the other representing the noisy loss distribution. Potential noisy samples are identified where the probability of being drawn from the noisy distribution exceeds a threshold $t$, determined by the intersection of the two distributions. Using these samples, the initial transition matrix for label correction $T \in [0,1]^{C \times C}$ is estimated. Noisy labels are flagged by a binary vector $b$
, where $b_n=1$ denotes a potential noisy label. During the training phase, we selectively apply $T$ to the noisy samples, while clean samples are trained with the standard cross-entropy loss.}
\label{fig:method}
\end{figure*}

The training stage shifts to a global perspective, where the information from the pre-training stage is used to improve the overall model training. Here, we incorporate loss corrections based on the noisy labels identified earlier. This correction is applied selectively, ensuring that the model learns more accurately from the clean data while adjusting its learning for noisy samples. This two-stage strategy allows us to maintain the integrity of the training process, leading to a more robust and reliable model that can generalize well despite the presence of noise in the labels.
An illustration of the entire method can be found in Figure~\ref{fig:method}.

\subsection{Pre-Training}

While various methods could be employed to differentiate noisy from clean labels, our pre-training procedure, inspired by Huang et al. \cite{huang2019o2unet}, focuses on tracking the loss values of each sample across epochs to identify noise-related patterns. This is done by applying a cyclic learning rate during training \cite{smith2017cyclical} and computing the loss values \( \ell \) for each sample \( x \). The cyclic learning rate helps escape sharp local minima and allows better separation of clean and noisy samples by amplifying the differences in their loss trajectories.

Then, to estimate which labels are noisy without prior knowledge about the contamination level, we introduce a \textbf{Loss Distribution Analysis}. Additionally, we implement \textbf{Transition Matrix Initialization} to optimize the subsequent training stage. These components work together to manage noisy labels and improve the overall model performance. 

\textbf{Loss Distribution Analysis}: 
Loss values, denoted as \( \boldsymbol{\ell} \), correspond to the per-sample negative log-likelihood (i.e., the cross-entropy loss), where each $\ell_n$ is defined by
\[
\ell_n = L(f_{\theta}(x_n), \tilde{y}_n).
\]
And the empirical probability density function (PDF) of these loss values observed during pre-training is denoted as \( g(\boldsymbol{\ell}) \). Specifically, we assume \( g(\boldsymbol{\ell}) \) follows a mixture of two Gaussian distributions \cite{muthen1999finite,reynolds2009gaussian}:
\[
g(\mathbf{\ell}) = \lambda \cdot \mathcal{N}(\ell; \mu_1, \sigma_1^2) + (1 - \lambda) \cdot\mathcal{N}(\ell; \mu_2, \sigma_2^2),
\]
where \( \mathcal{N}(\ell; \mu_1, \sigma_1^2) \) and \( \mathcal{N}(\ell; \mu_2, \sigma_2^2) \) are the probability density functions of the clean and contaminated (i.e., noisy) label loss distributions, respectively. Here, \( \mu_1 \) and \( \sigma_1^2 \) represent the mean and variance of the clean label loss distribution, while \( \mu_2 \) and \( \sigma_2^2 \) correspond to those of the contaminated label loss distribution. Finally, \( \lambda \) and \( 1 - \lambda \) are the mixing proportions of the two distributions, where \( \lambda \in (0,1) \).

All parameters, including the mixing proportion \( \lambda \), as well as the mean and variance of both Gaussian components (\( \mu_1, \sigma_1^2, \mu_2, \sigma_2^2 \)), are estimated using Maximum Likelihood via the Expectation-Maximization (EM) algorithm \cite{dempster1977maximum}. Once fitted, these distributions can be compared with the empirical distribution to estimate which labels are contaminated (i.e., noisy).

We define a threshold \( t \) on the loss values \( \ell \), such that a sample \( x \) with a loss \( \ell > t \) is classified as contaminated (i.e., noisy), while a sample with \( \ell \leq t \) is considered clean. This threshold is used to separate samples that are likely mislabeled from those that are likely correctly labeled. Then, the estimated true positive (TP), estimated false negative (FN), estimated false positive (FP), and estimated true negative (TN) rates can be expressed as shown in Table~\ref{tab:confusion_matrix}, where \( t \) is the threshold we want to learn. The algorithm is described in Algorithm 1.

\begin{table}[h]
    \centering
    \renewcommand{\arraystretch}{1.3}
    \setlength{\tabcolsep}{6pt} % Adjust column spacing
    \small % Use a smaller font to fit in a single column
    \caption{Confusion Matrix with Gaussian Distributions}
    \label{tab:confusion_matrix}
    \begin{tabular}{cc}
        \toprule
    {\textbf{TP}}
        &\textbf{FP} \\ 
        $\int_{-\infty}^{t} \lambda \cdot \mathcal{N}(x; \mu_1, \sigma_1^2) \, dx$ & 
        $\int_{-\infty}^{t} (1 - \lambda) \cdot \mathcal{N}(x; \mu_2, \sigma_2^2) \, dx$ \\
        \textbf{FN}
        & \textbf{TN} \\
        $\int_{t}^{\infty} \lambda\cdot  \mathcal{N}(x; \mu_1, \sigma_1^2) \, dx$ & 
        $\int_{t}^{\infty} (1 - \lambda) \cdot \mathcal{N}(x; \mu_2, \sigma_2^2) \, dx$ \\
        \bottomrule
    \end{tabular}
\end{table}

\begin{figure}[htb]
\begin{minipage}[b]{1.0\linewidth}
  \centering
  \centerline{\includegraphics[width=8.5cm]{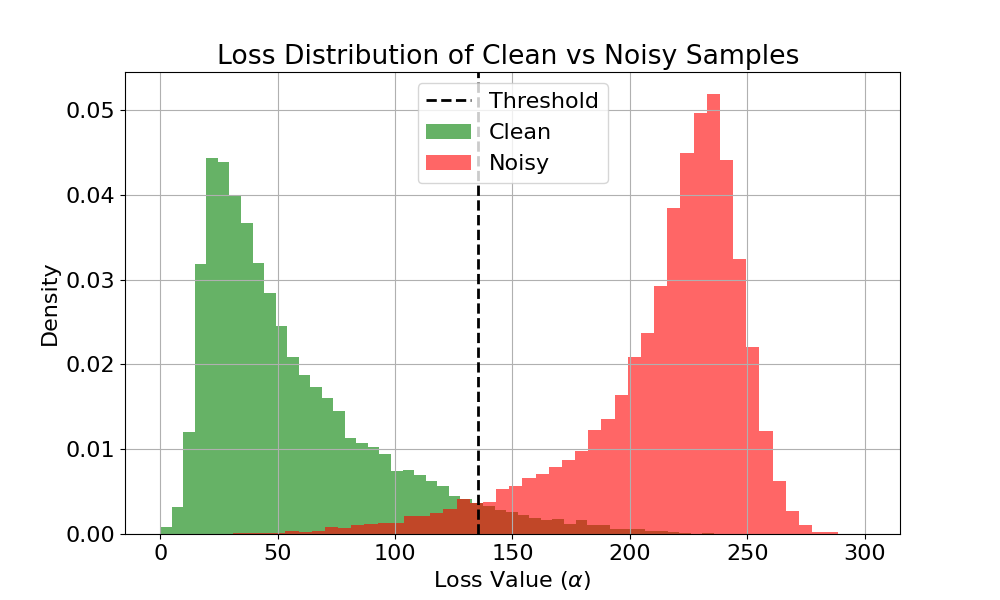}}
\end{minipage}
\caption{This figure illustrates the separation of clean and noisy examples based on their loss distributions under 50\% symmetric noise. It demonstrates how our method differentiates between clean and noisy labels through loss analysis, without any user input.}
\label{fig:gmm}
\end{figure}

We then seek to find the optimal threshold \( t \) that maximizes the balance between sensitivity (SEN) and specificity (SPC)  \cite{nouri2018optimized}:
\[
\max_{-\infty < t < \infty} \left( \frac{\text{SEN}(t) + \text{SPC}(t)}{2} \right),
\]
where sensitivity and specificity are defined as:
\[
\text{SEN}(t) = \frac{\text{TP}}{\text{TP} + \text{FN}}, \quad \text{SPC}(t) = \frac{\text{TN}}{\text{TN} + \text{FP}}.
\]
This criterion ensures that the selected threshold minimizes the probability of misclassification by balancing the rates of true positives and true negatives, as shown in Figure~\ref{fig:gmm}. By optimizing the average of sensitivity and specificity, we ensure that the threshold $t$ is neither too strict nor too lenient, leading to a well-balanced separation between clean and contaminated labels. Unlike methods that require prior knowledge of the noise rate or assume a fixed proportion of outliers, our approach estimates the contamination dynamically from the observed loss distribution. This makes our method more practical in real-world settings, where the actual noise level is often unknown.

Finally, using this optimized threshold, we create a binary vector \( b \), indicating each sample label as either noisy or clean. The binary vector \( b \) is defined as follows:
\[
b = \left[ b_1, b_2, \ldots, b_n \right] ,\\
\]
\[
\small
\text{where } b_n = \begin{cases} 
1 & \text{if loss } \ell_n \text{ exceeds threshold } t \text{ (noisy)}, \\
0 & \text{otherwise}.
\end{cases}
\]
\normalsize
\textbf{Transition Matrix Initialization}: Using all samples \( x_n \) such that \( b_n = 1 \), we estimate an initial transition matrix \( T \) by counting occurrences where the observed noisy label \( \tilde{y}_n = k \) differs from the classifier's prediction \( \hat{y}_n = l \), while the sample has been identified as noisy by the previous Loss Distribution Analysis:
\[
T_{l,k} = \sum^{N}_{n = 1} I(\tilde{y}_n = k, \; \hat{y}_n = l, \; b_n = 1), \hspace{2mm} \forall l,k \in \{1,2,\dots,C\}
\]
where \( I(\cdot) \) is an indicator function that equals 1 when its condition holds and 0 otherwise. Each row is then normalized to sum to 1. After normalization, \( T_{l,k} \) represents the empirical frequency of the classifier predicting label \( k \) given the label \( l \), only for suspected noisy labels. Since the initialization depends on classifier predictions, we assume the classifier is sufficiently accurate to provide a meaningful estimate of the noise structure.

\subsection{Model Training}
In this stage, we focus on enhancing the model's accuracy by addressing the challenges posed by noisy labels. As shown in Figure~\ref{fig:method}, we selectively apply corrections to the identified noisy data, ensuring that the model learns from the clean data without being misled by errors. This targeted approach helps the model to develop a more accurate and reliable understanding of the data, ultimately improving its overall performance.

\textbf{Selective Noise Correction}: During training, we modify the loss computation only for noisy samples while keeping the standard cross-entropy loss for clean samples.

For noisy samples (\( b_n = 1 \)), we first adjust the model predictions by incorporating the transition matrix. The cross-entropy loss is then given by:
\small % Adjust font size to fit within the column
\begin{align*}
L(\theta; x_n, \tilde{y}_n, b_n) =
\begin{cases}
    -\log P_{\theta}(\mathrm{Y} = \tilde{y}_n \mid x_n), & \text{if } b_n = 0, \\
    -\log \sum\limits_{l=1}^{C} T_{l,\tilde{y}_n} \cdot P_{\theta}(\mathrm{Y} = l \mid x_n), & \text{if } b_n = 1.
\end{cases}
\end{align*}
Here, \( T_{l,\tilde{y}_n} \) represents the transition probability from the true label \( l \) to the observed noisy label, and \( \theta \) represents the model parameters. The transition matrix \( T \) is continuously updated during training to adapt to evolving noise patterns in the dataset. By applying corrections only to noisy samples, we preserve the learning dynamics for clean data while improving robustness against label noise.

This hybrid approach combines noise detection with targeted correction, effectively leveraging noisy samples to improve model performance and enhance generalization.

Our approach builds on existing techniques but introduces key innovations for learning with noisy labels. Specifically, we use a \textbf{cyclical learning rate} \cite{smith2017cyclical} to track loss trajectories and detect noisy samples, whereas prior works used it mainly for optimization. We apply a GMM to separate noisy and clean samples, similar to DivideMix \cite{li2020dividemix}, but uniquely integrate it with cyclical learning for improved noise identification. Furthermore, we propose a novel \textbf{transition matrix initialization} based on this pre-training stage, which has not been done before. Finally, our \textbf{Selective Noise Correction} mechanism ensures that the transition matrix is applied only to noisy samples, avoiding unnecessary corrections to clean data- an aspect overlooked in prior noise correction methods. The algorithm is described in Algorithm 2.

\small
\begin{algorithm}
\caption{Pre-Training for Noise Detection and Transition Matrix Initialization}
\label{alg:pretraining}
\begin{algorithmic}[1]
\Require Dataset $\mathcal{\Tilde{D}} = \{(x_n, \tilde{y}_n)\}_{n=1}^N $, model $f_{\theta}$
\Ensure Updated model parameters $\theta$, Sample noise indicator vector $b$, transition matrix $T$
\State \textbf{Training:} Train $f_{\theta}$ on $\Tilde{\mathcal{D}}$ using a standard procedure with a scheduled cyclic learning rate, storing per-sample losses
\State \textbf{Noise Detection:} Fit a GMM to the loss distribution $g(\boldsymbol{\ell})$ 
% $g(\mathcal{A})$
\State Estimate threshold $t$ and identify noisy samples: $b_n \gets 1_{\ell_n > t},{n=1,\ldots,N}$
\State \textbf{Transition Matrix Initialization:} Initialize $T$ using $b$ and normalize row-wise
\State \Return $\theta$, $b, T$
\end{algorithmic}
\end{algorithm}
\normalsize

\small
\begin{algorithm}
\caption{Detect \& Correct}
\label{alg:ngtc}
\begin{algorithmic}[1]
\Require Dataset $\Tilde{\mathcal{D}} = \{(x_n, \tilde{y}_n)\}_{n=1}^N$, Model $f_{\theta}$, transition matrix $T$, noisy sample indicator $b$
\Ensure Updated model parameters $\theta$, corrected transition matrix $T$
\For{epoch in $N_{epochs}$}
    \For{each $(x_n, \tilde{y}_n)$ in $\Tilde{\mathcal{D}}$}
        \State Forward pass $f_\theta(x_n)$
        \State Compute $L(\theta; x_n, \tilde{y}_n, b_n)$
        \State Update transition matrix $T$
        \State Update model parameters $\theta$
    \EndFor
\EndFor
\State \Return $\theta$
\end{algorithmic}
\end{algorithm}
\normalsize

\section{Experiments}
\label{experiments}
\begin{table}[htbp]
\caption{Results on MNIST, CIFAR-10, CIFAR-100 for Pair-20\%  noise rate.}
\begin{center}
\resizebox{\columnwidth}{!}{ % This will resize the table to fit within the column width
\begin{tabular}{|c|c|c|c|}
\hline
\multirow{2}{*}{} & \multicolumn{1}{c|}{\textbf{MNIST}} & \multicolumn{1}{c|}{\textbf{CIFAR-10}} & \multicolumn{1}{c|}{\textbf{CIFAR-100}} \\ \cline{2-4}
 & \multicolumn{1}{c|}{Pair-20\%} & \multicolumn{1}{c|}{Pair-20\%} & \multicolumn{1}{c|}{Pair-20\%} \\ \hline
Decoupling & 96.93 $\pm$ 0.07 & 77.12 $\pm$ 0.30 & 40.12 $\pm$ 0.26 \\ \hline
MentorNet & 96.89 $\pm$ 0.04 & 77.42 $\pm$ 0.00 & 39.22 $\pm$ 0.47 \\ \hline
Co-teaching & 97.00 $\pm$ 0.06 & 80.65 $\pm$ 0.20 & 42.79 $\pm$ 0.79 \\ \hline
Forward & 98.84 $\pm$ 0.10 & 88.21 $\pm$ 0.48 & 56.12 $\pm$ 0.54 \\ \hline
T-Revision & 98.89 $\pm$ 0.08 & 90.33 $\pm$ 0.52 & 64.33 $\pm$ 0.49 \\ \hline
DMI & 98.84 $\pm$ 0.09 & 89.89 $\pm$ 0.45 & 59.56 $\pm$ 0.73 \\ \hline
Dual T & 98.86 $\pm$ 0.04 & 89.77 $\pm$ 0.25 & 67.21 $\pm$ 0.43 \\ \hline
VolMinNet & 99.01 $\pm$ 0.07 & 90.37 $\pm$ 0.30 & \textbf{68.45 $\pm$ 0.69} \\ \hline
D\&C-Net & \textbf{99.10 $\pm$ 0.09} & \textbf{91.86 $\pm$ 0.02} & 65.89 $\pm$ 0.48 \\ \hline
\end{tabular}
}
\label{tab:results_pair20}
\end{center}
\end{table}
\begin{table}[htbp]
\caption{Results on MNIST, CIFAR-10, CIFAR-100 for Sym-20\%  noise rate.}
\begin{center}
\resizebox{\columnwidth}{!}{ % This will resize the table to fit within the column width
\begin{tabular}{|c|c|c|c|}
\hline
\multirow{2}{*}{} & \multicolumn{1}{c|}{\textbf{MNIST}} & \multicolumn{1}{c|}{\textbf{CIFAR-10}} & \multicolumn{1}{c|}{\textbf{CIFAR-100}} \\ \cline{2-4}
 & \multicolumn{1}{c|}{Sym-20\%} & \multicolumn{1}{c|}{Sym-20\%} & \multicolumn{1}{c|}{Sym-20\%} \\ \hline
Decoupling & 97.04 $\pm$ 0.06 & 77.32 $\pm$ 0.35 & 41.92 $\pm$ 0.49 \\ \hline
MentorNet & 97.21 $\pm$ 0.06 & 81.35 $\pm$ 0.23 & 42.88 $\pm$ 0.41 \\ \hline
Co-teaching & 97.07 $\pm$ 0.10 & 82.27 $\pm$ 0.07 & 48.48 $\pm$ 0.66 \\ \hline
Forward & 98.60 $\pm$ 0.19 & 85.20 $\pm$ 0.80 & 54.90 $\pm$ 0.74 \\ \hline
T-Revision & 98.72 $\pm$ 0.10 & 87.95 $\pm$ 0.36 & 62.72 $\pm$ 0.69 \\ \hline
DMI & 98.70 $\pm$ 0.02 & 87.54 $\pm$ 0.20 & 62.65 $\pm$ 0.39 \\ \hline
Dual T & 98.43 $\pm$ 0.05 & 88.35 $\pm$ 0.33 & 62.16 $\pm$ 0.58 \\ \hline
VolMinNet & 98.74 $\pm$ 0.08 & 89.58 $\pm$ 0.26 & 64.94 $\pm$ 0.40 \\ \hline
D\&C-Net & \textbf{98.95 $\pm$ 0.03} & \textbf{90.61 $\pm$ 0.67} & \textbf{69.28 $\pm$ 0.08} \\ \hline
\end{tabular}
}
\label{tab:results_sym20}
\end{center}
\end{table}
\begin{table}[htbp]
\caption{Results on MNIST, CIFAR-10, CIFAR-100 for Sym-50\%  noise rate.}
\begin{center}
\resizebox{\columnwidth}{!}{ % This will resize the table to fit within the column width
\begin{tabular}{|c|c|c|c|}
\hline
\multirow{2}{*}{} & \multicolumn{1}{c|}{\textbf{MNIST}} & \multicolumn{1}{c|}{\textbf{CIFAR-10}} & \multicolumn{1}{c|}{\textbf{CIFAR-100}} \\ \cline{2-4}
 & \multicolumn{1}{c|}{Sym-50\%} & \multicolumn{1}{c|}{Sym-50\%} & \multicolumn{1}{c|}{Sym-50\%} \\ \hline
Decoupling & 94.58 $\pm$ 0.08 & 54.07 $\pm$ 0.46 & 22.63 $\pm$ 0.44 \\ \hline
MentorNet & 95.56 $\pm$ 0.15 & 73.47 $\pm$ 0.15 & 32.66 $\pm$ 0.40 \\ \hline
Co-teaching & 95.20 $\pm$ 0.23 & 75.55 $\pm$ 0.07 & 36.77 $\pm$ 0.52 \\ \hline
Forward & 97.77 $\pm$ 0.16 & 74.82 $\pm$ 0.78 & 41.85 $\pm$ 0.71 \\ \hline
T-Revision & 98.23 $\pm$ 0.10 & 80.01 $\pm$ 0.62 & 49.12 $\pm$ 0.22 \\ \hline
DMI & 98.12 $\pm$ 0.21 & 82.68 $\pm$ 0.22 & 52.42 $\pm$ 0.64 \\ \hline
Dual T & 98.15 $\pm$ 0.12 & 82.54 $\pm$ 0.19 & 52.49 $\pm$ 0.37 \\ \hline
VolMinNet & 98.23 $\pm$ 0.16 & 83.37 $\pm$ 0.25 & 53.89 $\pm$ 1.26 \\ \hline
D\&C-Net & \textbf{98.42 $\pm$ 0.03} & \textbf{87.30 $\pm$ 0.07} & \textbf{60.39 $\pm$ 0.08} \\ \hline
\end{tabular}
\label{tab:results_sym50}
}
\end{center}
\end{table}
\begin{table}[htbp]
\caption{Results on scRNA-seq PBMC cell-type annotation for Pair-20\%, Sym-20\%, and Sym-50\% noise rates.}
\begin{center}
\resizebox{\columnwidth}{!}{ % This will resize the table to fit within the column width
\begin{tabular}{|c|c|c|c|}
\hline
\multirow{2}{*}{} & \multicolumn{3}{c|}{\textbf{scRNA-seq PBMC cell-type annotation}} \\ \cline{2-4} 
\textbf{} & Pair-20\% & Sym-20\% & Sym-50\% \\ \hline
VolMinNet & 90.67 $\pm$ 0.77 & 90.85 $\pm$ 1.11 & 93.75 $\pm$ 0.08 \\ \hline
D\&C-Net & \textbf{93.83 $\pm$ 0.40} & \textbf{94.42 $\pm$ 0.29} & \textbf{93.93 $\pm$ 0.05} \\ \hline
\end{tabular}
}
\label{tab:results_scrna}
\end{center}
\end{table}

In this section, we evaluate the effectiveness of our proposed method using various datasets. We report the performance of our method in comparison to other state-of-the-art methods.
The code to reproduce our experiments is available at: \url{https://github.com/yuvalgrin/DetectAndCorrect-Net}.\\

\subsection{Datasets}
We performed experiments on the following datasets: MNIST \cite{lecun1998gradient}, CIFAR-10 \cite{krizhevsky2009learning}, CIFAR-100 \cite{krizhevsky2009learning} and an scRNA-seq PBMC dataset \cite{miao2020integrated}  \cite{hou2020systematic}: A dataset consisting of scRNA-seq cell count matrix data with 161,764 cells and 31 cell types.

\subsection{Experimental Setup}
For each dataset, we conducted experiments under varying levels of label noise to evaluate the robustness of our method. Specifically, we introduced 20\% and 50\% symmetric label noise (where labels are uniformly flipped to any other class) as defined in \cite{angluin1988learning}, as well as 20\%  pair label noise (where labels are flipped to a specific incorrect class) as defined in \cite{NEURIPS2018_a19744e2}. For both CIFAR-10 and CIFAR-100, we used 200 epochs for pre-training and 80 epochs for final training, while CIFAR-10 used ResNet-18 \cite{he2016deep} and CIFAR-100 used ResNet-34 \cite{he2016deep}.  For both MNIST and scRNA-seq PBMC cell-type annotation, we used 50 epochs for pre-training and 40 epochs for final training, while MNIST used LeNet and cell-type annotation used a three-layer fully connected model. We used a learning rate of \(10^{-2}\) for the model and \(10^{-4}\) for the transition matrix for all datasets and models.

\subsection{Results}
The results, summarized in Tables \ref{tab:results_pair20}, \ref{tab:results_sym20}, and \ref{tab:results_sym50}, demonstrate that our method consistently outperforms leading baselines across various noise levels. Specifically, our approach achieves higher classification accuracy, particularly in high-noise scenarios, underscoring its robustness and effectiveness in handling noisy labels. We compared our method with several established approaches, including Decoupling \cite{malach2017decoupling}, T-Reivision \cite{han2019deep}, MentorNet \cite{jiang2018mentornet}, Co-teaching \cite{NEURIPS2018_a19744e2}, Forward \cite{patrini2017making}, DMI \cite{xu2019l_dmi}, Dual T \cite{yao2020dual}, and VolMinNet \cite{Li2021ProvablyEL}.

\subsection{Single Cell RNA Sequencing Cell-Type Annotation}
Single-cell RNA sequencing (scRNA-seq) enables the examination of gene expression at the single-cell level, providing insights into cellular heterogeneity. Cell-type annotation, achieved by comparing gene expression profiles to known markers or reference datasets, is crucial for uncovering this heterogeneity. However, the low signal-to-noise ratio in scRNA-seq data, along with batch effects and dropout, makes label noise a significant challenge \cite{yuan2017challenges}, obscuring true cell-type distinctions and complicating accurate classification by learning algorithms. The results are summarized in Table~\ref{tab:results_scrna}, where our approach consistently achieves higher classification accuracy across all noise settings.

\textbf{Data preparation}: We utilized the Azimuth PBMC reference dataset \cite{miao2020integrated}. We then used totalVI \cite{gayoso2021totalvi} to integrate the cells into a 20-dimensional latent space to remove batch effects and facilitate the classifier training.

\subsection{Ablation}

We perform a detailed ablation study on D\&C-Net by first evaluating it without the selective noise correction and pre-training components on all datasets, which results in a notable drop in performance, with accuracy decreasing by an absolute \(\approx \mathbf{1\%-6\%}\) with an average of \(\approx \mathbf{2\%}\). We also examine the impact of not initializing the transition matrix \(T\), which leads to a smaller absolute decrease in the accuracy of \(\approx \mathbf{0.5\%}\) on the CIFAR-10 dataset. These results underscore the importance of both noise management and initialization strategies in achieving optimal performance.

\section{Discussion}
\label{discussion}
D\&C-Net integrates noisy label identification with selective loss correction, selectively applying the transition matrix to suspected noisy samples. Our experiments demonstrate that D\&C-Net consistently improves classification accuracy across diverse datasets, including MNIST, CIFAR-10, CIFAR-100, and scRNA-seq PBMC annotation. By preserving clean data integrity and applying targeted noise correction, D\&C-Net exhibits robustness, especially under high-noise conditions.

\subsection{Comparisons to Existing Methods}
Existing methods for learning with noisy labels generally fall into two categories: (1) \textit{loss correction techniques} that modify the loss function to account for noise, often via a noise transition matrix, and (2) \textit{data filtering approaches} that detect and remove noisy labels. D\&C-Net balances these two strategies by identifying noisy labels while preserving valuable training information through selective correction. Unlike DivideMix~\cite{li2020dividemix}, which treats all detected noisy labels as unlabeled data for semi-supervised learning, D\&C-Net refines learning by correcting mislabels rather than discarding them. This distinction ensures that our method better retains useful information from noisy data while mitigating the risk of confirmation bias introduced by mislabeled samples.

\subsection{Limitations and Future Directions}
One limitation of D\&C-Net is its reliance on accurate noisy label detection for transition matrix initialization. Errors in this step could lead to suboptimal noise correction, particularly in datasets with highly structured noise patterns or label-dependent noise. Future work could explore adaptive noise estimation techniques, such as iterative refinement of the transition matrix or integrating uncertainty estimation to better calibrate corrections.

Additionally, D\&C-Net could benefit from complementary techniques not explored in this paper, such as:
\begin{itemize}
    \item \textbf{Data Augmentation for Robustness:} Leveraging augmentation techniques, such as MixUp~\cite{zhang2018mixup} or RandAugment~\cite{cubuk2020randaugment}, could improve model generalization by reducing reliance on individual noisy samples.
    \item \textbf{Pseudo-Labeling for Semi-Supervised Learning:} Incorporating pseudo-labeling strategies~\cite{lee2013pseudo,eisenbergcoper} might enable D\&C-Net to refine its corrections by iteratively improving label quality.
    \item \textbf{Self-Supervised Pretraining:} Pretraining with self-supervised learning techniques, such as contrastive learning~\cite{chen2020simclr} or masked autoencoders~\cite{he2022mae}, could provide a stronger feature representation before applying D\&C-Net corrections.
    \item \textbf{Application to More Complex Noisy Scenarios:} Evaluating D\&C-Net on real-world datasets with domain shift \cite{roznerdomain} or multi-modal observations \cite{lindenbaum2015learning,lindenbaum2016multi} could further validate its effectiveness.
    \item \textbf{Use Case in Anomaly Detection:} In semi-supervised anomaly detection \cite{lindenbaum2024transductive,rozner2024anomaly}, labels are often noisy or unreliable. Evaluating our method under these conditions can offer valuable insights into its robustness. 
\end{itemize}

\subsection{Impact on Biological Data Analysis}
Our results on scRNA-seq PBMC annotation highlight the applicability of D\&C-Net beyond traditional image datasets. Label noise in single-cell data often arises from ambiguous cell-type boundaries and batch effects, making noise correction crucial for accurate annotation. Expanding D\&C-Net to integrate domain-specific biological priors, such as gene expression similarity metrics, could improve its ability to handle noise in biological datasets.

\subsection{Conclusion}
D\&C-Net presents a novel approach to learning with noisy labels by integrating noise detection with targeted loss correction. By applying the transition matrix selectively, it achieves a strong balance between robustness and data efficiency. Future work should focus on refining noise estimation, incorporating semi-supervised learning strategies, and extending the method to more diverse and complex datasets.

\section*{Acknowledgment}
We thank Immunai Inc. for inspiring this work, particularly in addressing noisy labels in cell-type classification using scRNA-seq datasets. We also thank Amitay Sicherman for his support and fruitful discussion.
\clearpage
\vfill
\pagebreak
\nolinenumbers
\printbibliography
\vfill
\pagebreak

\section{Supplementary Material}
\subsection{Statistical Significance Analysis}

We compared D\&C-Net with the strongest baseline, VolMinNet, using Welch's two-sample \textit{t}-test, which handles unequal variances and requires only summary statistics (\(\text{mean} \pm \text{SD}\) over three random seeds). 

Table~\ref{tab:ttest_results} reports the $p$-values: D\&C\text{-}Net is significantly better ($p<0.05$) in 3 of 6 settings, VolMinNet in 1 setting (CIFAR\text{-}100, Pair\text{-}20\%), and the remaining 2 settings show no significant difference.
\begin{table}[H]
\caption{Welch's \textit{t}-test comparing D\&C-Net to VolMinNet using summary statistics (mean $\pm$ SD over 3 seeds).}
\begin{center}
\resizebox{\columnwidth}{!}{
\begin{tabular}{|c|c|c|c|}
\hline
\textbf{Dataset / Noise} & \textbf{D\&C-Net} & \textbf{VolMinNet} & \textbf{p-value} \\ \hline
CIFAR-10 / Sym-50\% & 87.30 $\pm$ 0.07 & 83.37 $\pm$ 0.25 & $<$ 0.001 \\ \hline
CIFAR-100 / Sym-50\% & 60.39 $\pm$ 0.08 & 53.89 $\pm$ 1.26 & 0.012 \\ \hline
CIFAR-10 / Pair-20\% & 91.86 $\pm$ 0.02 & 90.37 $\pm$ 0.30 & 0.013 \\ \hline
CIFAR-10 / Sym-20\% & 90.61 $\pm$ 0.67 & 89.58 $\pm$ 0.26 & 0.100 \\ \hline
MNIST / Sym-50\% & 98.42 $\pm$ 0.03 & 98.23 $\pm$ 0.16 & 0.170 \\ \hline
CIFAR-100 / Pair-20\% & 65.89 $\pm$ 0.48 & 68.45 $\pm$ 0.69 & 0.0085 \\ \hline
\end{tabular}
}
\label{tab:ttest_results}
\end{center}
\end{table}

\subsection{Computational Cost Comparison}

D\&C-Net introduces a single pre-training phase with a cyclical learning rate, adding roughly \(1.5\text{--}2\times\) the wall-clock time of a plain cross-entropy run while keeping memory usage virtually unchanged. This overhead is comparable to other two-stage or dual-network approaches such as Co-teaching and DivideMix, but higher than single-stage transition-matrix methods like Forward, T-Revision, and VolMinNet.

\end{document}